\pdfoutput=1
\documentclass[lettersize,journal]{IEEEtran}

\usepackage[utf8]{inputenc}
\usepackage[T1]{fontenc}
\usepackage{graphicx}
\usepackage{amsmath,amssymb,amsfonts}
\usepackage{booktabs}
\usepackage{multirow}
\usepackage{makecell}
\usepackage{colortbl}
\definecolor{ourshade}{rgb}{0.88,0.93,1.0}     % light blue: our method row
\definecolor{avgshade}{rgb}{0.92,0.92,0.92}    % light grey: summary columns
\usepackage{array}
\usepackage{stfloats}   % IEEE sample: allow bottom floats in 2-column
\usepackage{xcolor}
\usepackage{url}
\usepackage{cite}     % IEEE sample: compressed numeric citation lists
\usepackage{algorithm}
\usepackage{algorithmic}
\usepackage{tikz}
\usetikzlibrary{arrows.meta,positioning,shapes.geometric,fit,backgrounds,calc}
\usepackage{textcomp}
\usepackage{xspace}
\usepackage{caption}
\usepackage[hidelinks]{hyperref}
\usepackage[capitalize]{cleveref}

\newcommand{\sysname}{CADET\xspace}   % the framework (plain caps, no small-caps styling)
\newcommand{\pcr}{PCR\xspace}         % Physics-grounded Causal Reliance score
\newcommand{\tcm}{TCM\xspace}         % Test-time Causal Masking

\newcommand{\dox}[1]{\mathrm{do}(#1)}

\definecolor{cdInk}{HTML}{14171B}      % primary text
\definecolor{cdSlate}{HTML}{5C6672}    % secondary text, neutral strokes
\definecolor{cdLine}{HTML}{C3C9D0}     % hairlines, panel borders
\definecolor{cdPanel}{HTML}{FCFCFD}    % panel background
\definecolor{cdWell}{HTML}{F1F3F5}     % inset background (scene, charts)

\definecolor{cdBlue}{HTML}{2A5D8F}     % CADET accent
\definecolor{cdBlueF}{HTML}{E7EEF6}
\definecolor{cdGreen}{HTML}{226B54}    % causal / protected
\definecolor{cdGreenF}{HTML}{E4F0EA}
\definecolor{cdAmber}{HTML}{C96A2A}    % spurious / flagged
\definecolor{cdAmberF}{HTML}{F9E1CE}
\definecolor{cdGrey}{HTML}{7E8894}     % frozen planner
\definecolor{cdGreyF}{HTML}{EDEFF2}

\providecommand{\cdnohyph}{\hyphenpenalty=10000\exhyphenpenalty=10000\relax}
\providecommand{\cdtiny}{\fontsize{6}{6.6}\selectfont}

\providecommand{\sysname}{CADET\xspace}
\providecommand{\pcr}{PCR\xspace}
\providecommand{\tcm}{TCM\xspace}

\tikzset{
  cdfig/.style={
    x=1.0574cm, y=1.0574cm, font=\scriptsize, line width=0.5pt,
    text=cdInk, >={Stealth[round,length=3.4pt,width=2.6pt]}},
  panel/.style={draw=cdLine, fill=cdPanel, rounded corners=2.6pt,
    line width=0.5pt},
  well/.style={draw=cdLine, fill=cdWell, rounded corners=2pt,
    line width=0.4pt},
  card/.style={draw=cdLine, fill=white, rounded corners=2pt,
    line width=0.4pt},
  ptitle/.style={font=\footnotesize\bfseries, inner sep=0pt, anchor=west},
  psub/.style={font=\cdtiny, text=cdSlate, inner sep=0pt, anchor=west},
  lab/.style={font=\cdtiny, text=cdInk, inner sep=0pt},
  note/.style={font=\cdtiny, text=cdSlate, inner sep=0pt, align=left,
    execute at begin node=\cdnohyph},
  chip/.style={rounded corners=1.6pt, inner xsep=2.6pt, inner ysep=1.6pt,
    font=\cdtiny, line width=0.4pt},
  chipG/.style={chip, draw=cdGreen, fill=cdGreenF, text=cdGreen!62!black},
  chipA/.style={chip, draw=cdAmber, fill=cdAmberF, text=cdAmber!70!black},
  chipB/.style={chip, draw=cdBlue,  fill=cdBlueF,  text=cdBlue!72!black},
  flow/.style={->, draw=cdSlate, line width=0.6pt},
  hair/.style={draw=cdLine, line width=0.4pt},
  probe/.style={->, draw=cdSlate!70, line width=0.45pt, dash pattern=on 1.4pt off 1.1pt},
}

\tikzset{
  pics/snow/.style={code={
    \foreach \a in {90,150,210}{
      \draw[line width=0.32pt, draw=cdBlue!75]
        (\a:1.75pt) -- ($(0,0)-(\a:1.75pt)$);}
    \foreach \a in {90,210,330}{
      \draw[line width=0.28pt, draw=cdBlue!75]
        (\a:1.75pt) -- ++($(\a+40:0.75pt)$)
        (\a:1.75pt) -- ++($(\a-40:0.75pt)$);}
  }}
}

\begin{document}

\title{\sysname: Physics-Grounded Causal Auditing of End-to-End
Driving Planners}

\author{%
Zikun~Guo\textsuperscript{1,2},\
Minglan~Chen\textsuperscript{1,2},\
Jinyou~Zhai\textsuperscript{1,2}, and\
Rongjin~Zou\textsuperscript{1,2}\\[3pt]
{\normalsize\textsuperscript{1}Geely Central Research Institute}\\[1pt]
{\normalsize\textsuperscript{2}School of Intelligent Driving, Xinwei
Institute of Artificial Intelligence}\\[1pt]
{\small\itshape gzk798412226@gmail.com, myangel1972@sina.com,
a710644534@gmail.com, 330494275@qq.com}%
}

\markboth{Preprint. Under review.}%
{Guo \MakeLowercase{\textit{et al.}}: Physics-Grounded Causal Auditing of Driving Planners}

\maketitle

\begin{abstract}
End-to-end (E2E) autonomous-driving planners trained by imitation can exploit
spurious correlations: they link scene elements that merely co-occur with
expert actions, such as a roadside object or a building facade, to driving
decisions, instead of the variables that causally determine them. Such causal
confusion is hard to detect, because the prevailing open-loop metrics, L2
displacement and collision rate, are dominated by ego status and say little
about whether a planner depends on spurious cues. Existing remedies based on
causal-intervention training require retraining and cannot audit an already
deployed planner. We present \sysname, a training-free framework that
measures spurious reliance in a pretrained E2E planner and localizes it
to individual planning queries, with no parameter update. \sysname has three
components. First, a Physics-grounded Causal Reliance score (\pcr) identifies
planning queries that the model depends on but that cannot physically
influence the decision. It fuses perturbation-based model influence with a
physics-geometric prior (time-to-collision, drivable-corridor relevance, and
object class). The prior is read off scene geometry, so it is an external
anchor that is independent of the training distribution. Second, a
counterfactual robustness benchmark with three perturbation families
(spurious, causal-link, and distribution-shift) and three metrics (CSI, CRI,
and CCS). We show that the natural stability probe is degenerate: it is
satisfied by construction whenever it selects its targets with the same score
the repair masks by. We give two probes that are decoupled from it. Third,
Test-time Causal Masking (\tcm) suppresses the flagged queries at inference,
which shows that the located dependencies are removable without training. On a
controlled benchmark in which the status of every object is fixed by
construction, \pcr localizes spurious reliance at an F1 of $0.92$ against
$0.50$ for the next-best training-free criterion, and it alone reaches that
precision while masking under $13\%$ of queries and keeping causal response
intact. On the pretrained SparseDrive planner evaluated on nuScenes, the audit
measures a causal-to-spurious per-agent influence ratio of $9.5$, yet in
individual frames the most influential physically irrelevant agent displaces
the plan as strongly as a genuine causal agent. Suppressing those agents moves
the official open-loop metrics by under $0.1$\,m, which shows that
displacement error does not register a change in causal reliance and is
therefore unsuited to measuring it.

\end{abstract}


\section{Introduction}
\label{sec:intro}

\begin{figure*}[t]
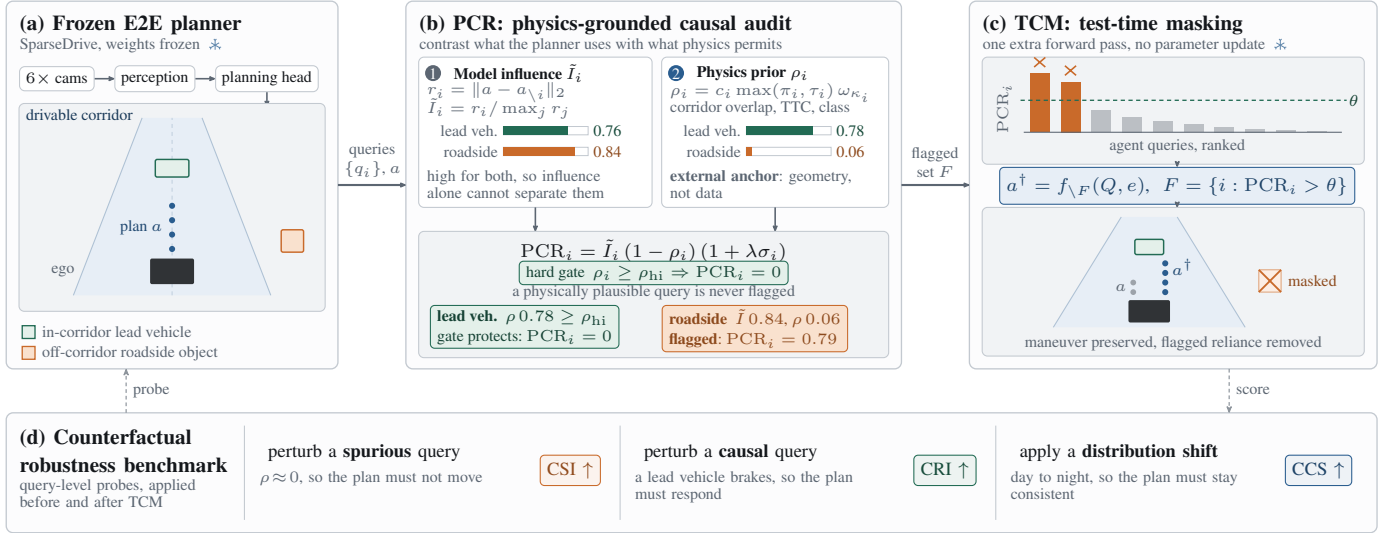

\centering
% =====================================================================
%  Fig. 1: CADET overview.
%  Assembles four TikZ panels drawn in their own local frames:
%    (a) frozen planner   (b) PCR audit   (c) TCM   (d) benchmark plate
%  Requires figures/fig1/style.tex in the preamble.
% =====================================================================
\begin{tikzpicture}[cdfig]
% ---- canvas layout (cm) ---------------------------------------------
\def\Wtot{17.15}
\def\pRowy{2.05}          % baseline of the pipeline row
\def\pAx{0.00}            % panel (a), width 4.15
\def\pBx{5.00}            % panel (b), width 6.20
\def\pCx{12.05}           % panel (c), width 5.10

\input{figures/fig1/panel_a}
\input{figures/fig1/panel_b}
\input{figures/fig1/panel_c}
\input{figures/fig1/panel_d}

% ---- stage-to-stage flow --------------------------------------------
\draw[flow, line width=0.8pt] (4.15,4.35) -- (5.00,4.35);
\node[note, anchor=south, align=center, text=cdSlate] at (4.58,4.44)
  {queries\\[0.5pt]$\{q_i\}$, $a$};
\draw[flow, line width=0.8pt] (11.20,4.35) -- (12.05,4.35);
\node[note, anchor=south, align=center, text=cdSlate] at (11.62,4.44)
  {flagged\\[0.5pt]set $F$};

% ---- the benchmark probes the pipeline it sits under -----------------
\draw[probe] (1.50,1.50) -- (1.50,2.05);
\node[note, anchor=west, text=cdSlate] at (1.58,1.78) {probe};
\draw[probe] (15.30,2.05) -- (15.30,1.50);
\node[note, anchor=west, text=cdSlate] at (15.38,1.78) {score};
\end{tikzpicture}
\caption{Overview of \sysname. \textbf{(a)} A pretrained end-to-end
planner is run unmodified and emits vectorized agent queries together with
an ego plan. \textbf{(b)} \pcr audits each query by contrasting two
signals: the model influence $\tilde I_i$, obtained by ablating the query
and measuring the induced plan change, and a physics-geometric prior
$\rho_i$ computed from drivable-corridor overlap, time-to-collision, and
object class. The planner relies on the in-corridor lead vehicle and on the
off-corridor roadside object to a comparable degree, so influence alone
cannot separate them. Because $\rho$ is read off scene geometry rather than
training statistics, it acts as an external anchor, and a hard gate on
$\rho$ guarantees that a physically plausible query is never flagged. The
values shown are the per-category means measured on SpurGen
(\cref{tab:spurgen}). \textbf{(c)} \tcm suppresses the flagged set at
inference and re-plans, which removes the spurious reliance while
preserving the maneuver. \textbf{(d)} A counterfactual benchmark probes the
same frozen planner along three perturbation families and scores it with
CSI, CRI, and CCS. The pipeline requires no parameter update and runs
inference-only on a single 16\,GB GPU.}
\label{fig:pipeline}
\end{figure*}

% --- P1: phenomenon ---
\IEEEPARstart{E}{nd-to-end} (E2E) autonomous driving has become the dominant
paradigm since UniAD~\cite{uniad}, with VAD~\cite{vad}, SparseDrive~\cite{sparsedrive},
and PARA-Drive~\cite{paradrive} further improving planning accuracy and
efficiency. These systems are trained by imitation on large logged
datasets, which introduces a structural weakness: imitation learning is
non-causal and can exploit spurious correlations, namely features that
co-occur with expert actions in the data but do not cause them~\cite{dehaan}. A planner may, for instance, learn to decelerate near a
particular building facade or roadside object because such decelerations
were frequent in the logs, while disregarding the agent that genuinely
warrants braking. This causal confusion is a principal obstacle to advancing E2E
driving from L2 to L3/L4 reliability.

% --- P2: the measurement gap (our wedge) ---
Two factors make this failure mode hard to address. First, it is hard to
observe. Evaluation still relies mainly on open-loop L2 displacement and
collision rate on nuScenes. These metrics are dominated by ego status and
change little when the perception input is removed~\cite{egostatus}, so a low
L2 error says little about whether a planner relies on spurious cues. Second,
it is costly to repair. Recent causal-intervention methods such as
CausalVAD~\cite{causalvad} reduce confounding by retraining the planner with
a sparse intervention module. Retraining is expensive, and it gives no way to
audit a planner that has already been trained and deployed. What is missing is
a training-free tool that determines whether, where, and how much a given
pretrained planner relies on spurious correlations.

% --- P3: our idea ---
We address this gap with \sysname (Causal Auditing and Deconfounding at
Test-time), a training-free framework whose primary object is
measurement: given a frozen planner, it determines which of the
planner's own queries carry reliance that physics cannot justify, and it
does so without parameter updates (\cref{fig:pipeline}). Our central observation is that true causes cannot be
distinguished from globally spurious correlates on the basis of
training-data statistics alone: when a spurious co-occurrence holds across
all environments, observational signals such as perturbation sensitivity and
cross-environment invariance both fail under the same faithfulness
assumption~\cite{irm}. We therefore introduce an external anchor: a
physics-geometric prior derived from the perception module's own outputs
(3D boxes, velocity, class, and detection confidence). This prior determines
whether an object is capable of influencing the decision at all, through
time-to-collision, drivable-corridor relevance, and dynamic or static state,
independently of the data distribution. A query on which the planner relies
heavily, yet which the physics prior deems irrelevant, is by construction a
spurious dependency.

% --- P4: contributions ---
\noindent\textbf{Contributions.}
\begin{itemize}
  \item \textbf{Physics-grounded causal audit (\pcr).} A training-free,
  model-agnostic score that flags per-query spurious reliance by fusing
  perturbation-based model influence with a physics-geometric prior, which
  acts as a hard external arbiter, together with a cross-environment
  stability term. It requires only forward passes on a pretrained planner
  (\cref{sec:method}).
  \item \textbf{Counterfactual robustness benchmark, with a probe design
  that is not self-fulfilling.} A standardized protocol with three
  perturbation families (spurious, causal-link, and distribution-shift) and
  three metrics, the Causal Stability Index (CSI), Causal Response Index
  (CRI), and Causal Consistency Score (CCS). We show that the obvious way to
  build the stability probe is degenerate, since selecting its targets by
  the physics score that the repair also masks by makes the probe a null
  operation, and we replace it with a ground-truth probe and an insertion
  probe that no masking rule can satisfy by deletion
  (\cref{sec:method:bench}, \cref{sec:res:probe}).
  \item \textbf{A demonstration that the located reliance is removable
  (\tcm).} Test-time Causal Masking suppresses the flagged queries at
  inference, realizing an approximate $\dox{\cdot}$ intervention at
  deployment with no training. \tcm shows that the audit localizes an
  actionable set: the flagged dependencies can be removed at test time while
  the maneuver is preserved (\cref{sec:res:bench}).
  \item \textbf{Reproducibility on commodity hardware.} The full pipeline
  is inference-only and runs on a single 16\,GB GPU. We release the code,
  the benchmark, and the audit toolkit.
\end{itemize}

% --- P5: scope ---
\sysname is complementary to existing causal methods. Unlike
CausalVAD~\cite{causalvad} and Beyond Patterns~\cite{beyondpatterns},
which retrain models for planning or prediction, and unlike generative
world-model approaches, which synthesize counterfactual scenarios for
closed-loop testing, \sysname neither trains nor generates: it instruments
existing planners and is therefore directly applicable to deployed systems. The remainder of the paper reviews related
work (\cref{sec:related}), formalizes the causal setting
(\cref{sec:prelim}), details \sysname (\cref{sec:method}), describes the
protocol (\cref{sec:experiments}), reports the audit (\cref{sec:results}),
and discusses implications and scope (\cref{sec:discussion},
\cref{sec:conclusion}).

\section{Related Work}
\label{sec:related}

\subsection{End-to-End Driving Planners}
Deep learning now covers tasks as varied as efficient transformer
architectures~\cite{guoclustertransformer},
model compression for question answering~\cite{guoidsextract}, generative
modeling~\cite{guoclustergan}, speech enhancement~\cite{guospeech}, visual
recognition~\cite{guocropvit}, reinforcement learning for combinatorial
optimization~\cite{guodytrl}, and multi-robot coordination and
scheduling~\cite{guoantarcticga,guoivec}. Within this progress,
vision-centric driving models have grown into end-to-end (E2E) systems
that map raw sensor input directly to a plan. UniAD~\cite{uniad}
introduced a planning-oriented architecture that unifies perception,
prediction, and planning; VAD~\cite{vad} replaced dense rasterization with
a vectorized scene representation; and SparseDrive~\cite{sparsedrive} and
PARA-Drive~\cite{paradrive} improved efficiency through sparse queries and
parallelized auxiliary tasks. These planners are trained by imitation and
are evaluated mainly by open-loop L2 displacement and collision rate on
nuScenes. Li~et~al.~\cite{egostatus} show that these metrics are dominated by
ego status: removing the perception input leaves the reported values almost
unchanged. A low open-loop error therefore says little about whether a
planner reasons over the scene or exploits spurious regularities. This is
what the measure of causal reliance developed here is for.

\subsection{Causal Confusion and Deconfounding in Driving}
The non-causal nature of behavioral cloning was established by de
Haan~et~al.~\cite{dehaan}, who showed that imitators rely on effects
rather than causes, such as a brake-indicator light, so that additional
observation can yield worse policies under distribution shift. In driving,
CausalVAD~\cite{causalvad} is the closest precursor to our work: it
instantiates do-calculus as a sparse causal intervention scheme, builds a
dictionary of context prototypes, and performs backdoor adjustment on
vectorized queries, achieving state-of-the-art open-loop planning. Beyond
Patterns~\cite{beyondpatterns} applies a diffusion-based backdoor
adjustment to map features for trajectory prediction. Closer to the
imitation-planning literature, PLUTO~\cite{pluto} and
PlanTF~\cite{plantf} curb spurious reliance with training-time data
augmentation, dropping the leading agent or masking the ego state so the
planner cannot lean on them. All of these methods retrain the planner, and each identifies confounders
from learned features or predefined attributes, without an external,
distribution-independent criterion for causal status. \sysname differs on
both counts: it operates on a frozen planner without parameter updates, and
it anchors the spurious-versus-causal distinction in a physics-geometric
prior rather than in data statistics. This covers the
global-spurious-correlation case in which feature-based identification fails.
We position \sysname against the families of work relevant to deployed
systems: causal-intervention deconfounders
(CausalVAD~\cite{causalvad}, Beyond Patterns~\cite{beyondpatterns},
PLUTO~\cite{pluto}, PlanTF~\cite{plantf}), the behavioral-cloning analyses
that first exposed causal confusion~\cite{dehaan}, invariance
learning~\cite{irm,icp}, perturbation attribution~\cite{occlusion}, and
generative or closed-loop evaluation~\cite{bench2drive,navsim}. Across these
families, \sysname is the only approach that operates on a frozen planner,
repairs it at inference, and grounds the causal decision in an external,
distribution-independent physics anchor. The training-free modules that are
applicable to a frozen planner are compared quantitatively in
\cref{tab:comparison}. The retraining-based and other test-time methods are
positioned here instead, because they cannot be applied to a frozen model as
a flagging module. End-to-end planners such as UniAD, VAD, and
SparseDrive are not competitors here but the subjects of the audit:
\sysname instruments a given planner, and its interface is planner-agnostic
(\cref{sec:exp:nusc}).

\subsection{Counterfactual Evaluation and Robustness}
A complementary line of work evaluates robustness through interventions.
Generative world models synthesize safety-critical counterfactual
scenarios for closed-loop testing, and Bench2Drive~\cite{bench2drive} and
NAVSIM~\cite{navsim} provide closed-loop and non-reactive simulation
benchmarks. Both are motivated by the gap between open-loop L2 and
closed-loop behavior. Invariance-based methods such as invariant risk
minimization~\cite{irm} seek predictors that are stable across
environments, but they rest on a faithfulness assumption that fails
precisely when a spurious correlate is invariant across all observed
environments. Our counterfactual benchmark targets a different goal, namely measuring how
much a given pretrained planner relies on spurious versus causal factors. It
uses inexpensive, deterministic query-level perturbations rather than a
generative simulator, so it is reproducible on commodity hardware.

\paragraph{Plug-and-play, test-time methods.} Because \sysname operates on
a frozen planner at inference, its true peers are plug-and-play,
training-free methods applied at test time. Recent work in this vein
improves a deployed planner without retraining: TOAD~\cite{toad} optimizes
the output trajectory at test time with a sampling search, and
Centaur~\cite{centaur} adapts the planner online through test-time
training. These methods target overall driving quality rather than per-agent spurious
reliance. TOAD's trajectory search and Centaur's online updates are
orthogonal to, and composable with, our masking operator. The criteria
directly comparable to \sysname are those that, like it, flag or suppress
agent queries at inference: perturbation attribution (occlusion
analysis)~\cite{occlusion} ranks queries by the output change they cause;
a training-free invariance test in the spirit of ICP/IRM~\cite{icp,irm}
flags queries whose influence is unstable across environments; and simple
detector-driven heuristics (masking low-confidence or randomly selected
queries) require no signal at all. All of these criteria can drive the same
test-time masking operator, and we therefore benchmark against them across
several planners in \cref{tab:comparison}. We find that the influence- and invariance-based
criteria improve stability only by masking genuinely causal agents as well.
\sysname differs from this family through the external physics anchor, which
is what lets it suppress spurious reliance while keeping causal response.

\subsection{Scene Perception and Structural Priors}
The physics prior in \sysname is computed from the perception module's own
outputs, so understanding street scenes from imagery is an enabling capability,
recently advanced by combining large language models with street-view
data~\cite{guostreet}. Using a structural or physical prior to guide
decision-making follows the same idea as its use in balanced and scalable
multi-robot path planning~\cite{guopath}. \sysname
applies the same principle to E2E driving, using a kinematic prior as an
external anchor against spurious reliance.

\section{Problem Formulation}
\label{sec:prelim}

\subsection{A Structural Causal Model for Planning}
\label{sec:prelim:scm}
\begin{figure}[t]
\centering
\begin{tikzpicture}[
  every node/.style={draw, circle, minimum size=8mm, inner sep=1pt,
    font=\small},
  >={Stealth[round]}, node distance=10mm and 12mm]
  \node (C) {$C$};
  \node (S) [below left=of C] {$S$};
  \node (Y) [below right=of C] {$Y$};
  \draw[->] (C) -- (S);
  \draw[->] (S) -- (Y);
  \draw[->, dashed] (C) -- (Y);
\end{tikzpicture}
\caption{Structural causal model of end-to-end planning. Scene context
$C$ confounds perception state $S$ and plan $Y$ via the backdoor path
$Y\!\leftarrow\! C\!\rightarrow\! S$ (dashed). An imitation-trained planner
fits the confounded $P(Y\mid S)$, whereas \sysname targets the deconfounded
$P(Y\mid \mathrm{do}(S))$.}
\label{fig:scm}
\end{figure}
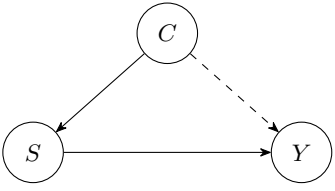

We model an end-to-end planner with a structural causal model (SCM)
$\mathcal{M}=(\mathbf{U},\mathbf{V},\mathcal{F},P(\mathbf{U}))$. The
endogenous variables $\mathbf{V}$ comprise a latent \emph{scene context}
$C$ (weather, road geometry, traffic density, and data-collection
biases), the \emph{perception state} $S$ produced by the perception
stack, and the \emph{plan} $Y$ (the future ego trajectory). The exogenous
$\mathbf{U}$ collect unobserved environmental noise. The relevant causal
structure (\cref{fig:scm}) is
\begin{equation}
  C \rightarrow S, \qquad C \rightarrow Y, \qquad S \rightarrow Y,
  \label{eq:graph}
\end{equation}
so that $C$ is a \emph{confounder} opening a backdoor path
$Y \leftarrow C \rightarrow S$. An imitation-trained planner fits the
observational $P(Y\mid S)$, which is biased by this backdoor path. The
quantity of interest is instead the interventional $P(Y\mid \dox{S})$. Under the backdoor criterion with adjustment set
$C$,
\begin{equation}
  P(Y\mid \dox{S}) \;=\; \sum_{c} P(Y\mid S,\,C{=}c)\,P(C{=}c).
  \label{eq:backdoor}
\end{equation}
Eq.~\eqref{eq:backdoor} is the objective that training-time methods such as
CausalVAD~\cite{causalvad} approximate through retraining. Our objective
differs: we audit and approximate this adjustment at test time on a frozen
planner. The tendency of modern models to exploit
context-induced spurious cues rather than causal ones is not specific to
driving. It recurs across modalities, for instance as sycophancy in video
and medical vision--language models~\cite{guosyc1,guosyc2}, which motivates
a general, model-agnostic treatment.

\subsection{Query-Level Notation}
\label{sec:prelim:query}
Modern planners~\cite{uniad,vad,sparsedrive} expose a set of vectorized
queries $Q=\{q_1,\dots,q_N\}$, each $q_i$ encoding a perceived entity
(agent or map element) together with physical attributes
$\phi_i=(\mathbf{b}_i,\mathbf{v}_i,\kappa_i,c_i)$: 3D box, velocity,
class, and detection confidence. With ego state $e$, the planner emits a
trajectory $a=f(Q,e)\in\mathbb{R}^{T\times2}$ over a horizon $T$. We
define a \emph{removal operator} $m_i$ that replaces $q_i$ by a baseline
(mean query or null token),
\begin{equation}
  f_{\setminus i}(Q,e) \;\triangleq\; f\big(m_i(Q),\,e\big),
  \label{eq:remove}
\end{equation}
and a set version $f_{\setminus F}$ for a subset $F\subseteq\{1,\dots,N\}$.

\subsection{Spurious vs.\ Causal Reliance}
\label{sec:prelim:reliance}
The \emph{reliance} of the plan on query $i$ is the decision change under
its removal,
\begin{equation}
  r_i \;=\; \big\| f(Q,e) - f_{\setminus i}(Q,e) \big\|_2 .
  \label{eq:reliance}
\end{equation}
A query is \emph{causally relevant} if the entity it represents can
physically affect the ego decision, such as a lead vehicle or a crossing
pedestrian. It is \emph{spurious} if $r_i$ is large yet the entity cannot
physically influence that decision, as in the case of a distant static
object off the ego path. The central difficulty is that observational
signals alone cannot separate the two: when a spurious co-occurrence holds
across all training environments (a global spurious correlation), both a
high reliance $r_i$ and a low cross-environment variance of $r_i$ wrongly
indicate causality. This is the failure of the faithfulness assumption that
limits invariance-based methods~\cite{irm}. We therefore introduce an external,
distribution-independent anchor, namely a physics-geometric prior, in
\cref{sec:method}.

\section{The \sysname Framework}
\label{sec:method}

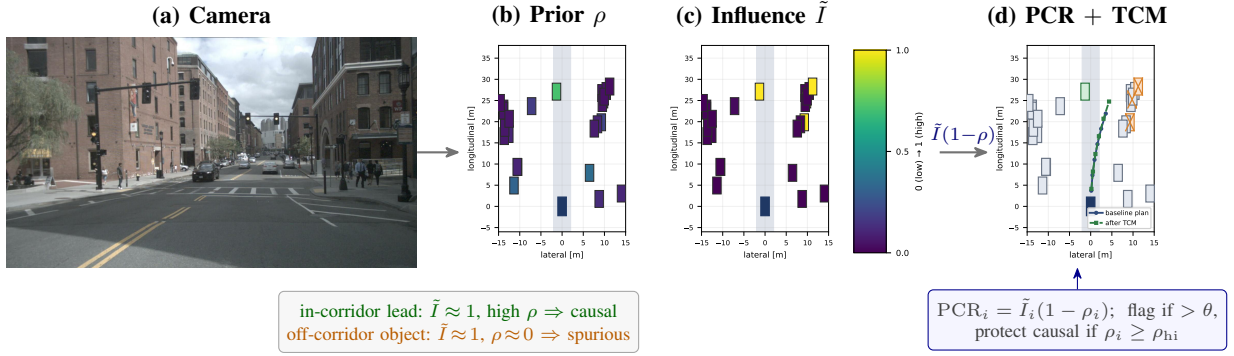
\begin{figure*}[t]
\centering
\begin{tikzpicture}[font=\footnotesize, >={Stealth[round]},
  panel/.style={inner sep=0pt},
  ttl/.style={font=\small\bfseries, inner sep=1.5pt},
  call/.style={font=\scriptsize, align=center, text=black!75,
    rounded corners=2pt}]
\node[panel] (a) {\includegraphics[height=3.05cm]{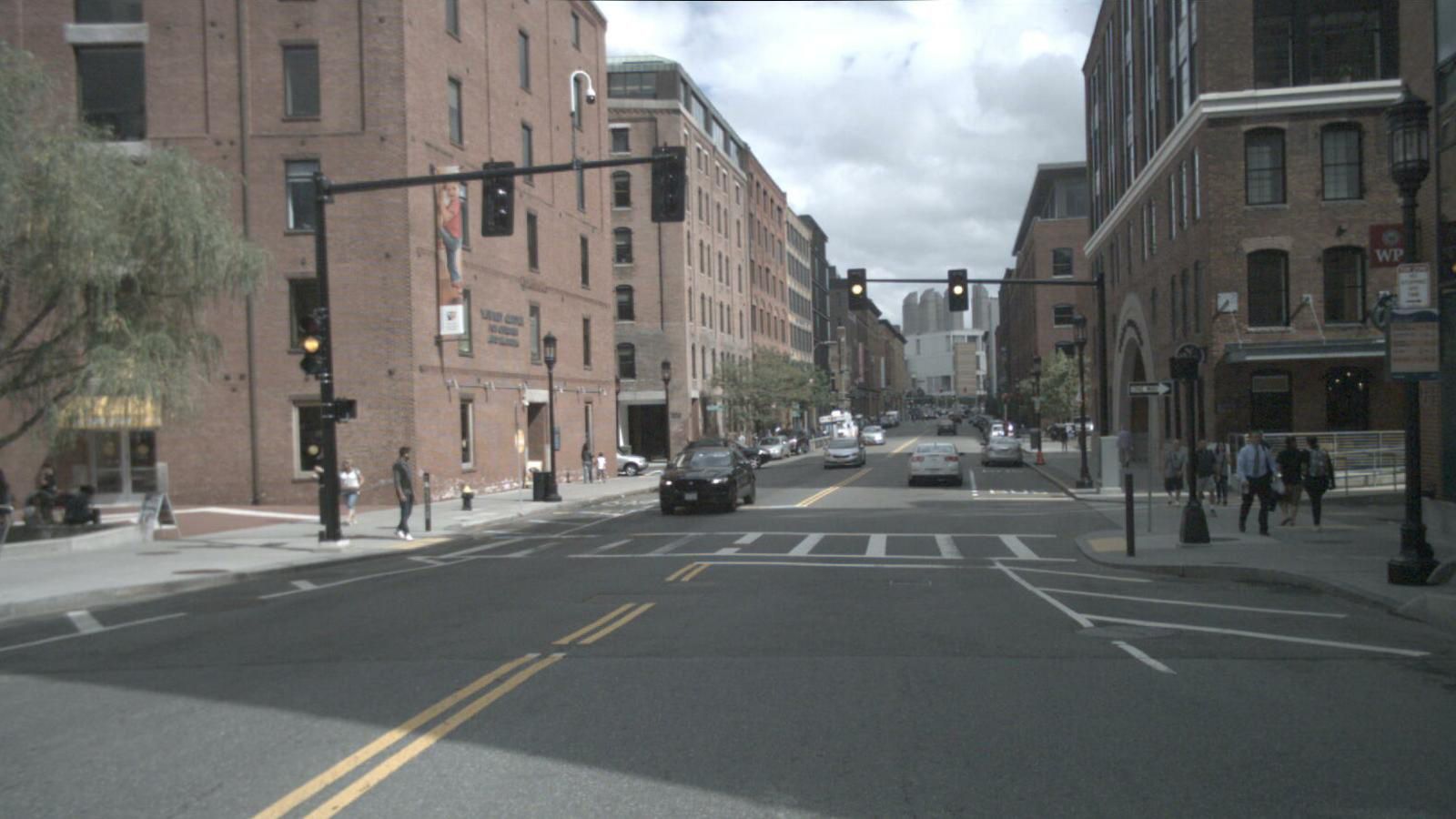}};
\node[panel, right=0.55cm of a] (b)
  {\includegraphics[height=3.05cm]{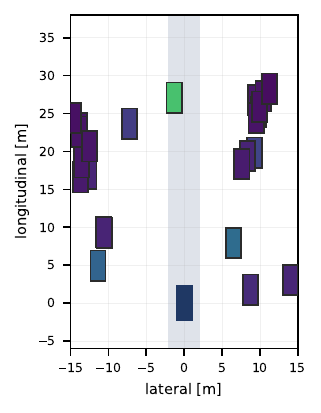}};
\node[panel, right=0.30cm of b] (c)
  {\includegraphics[height=3.05cm]{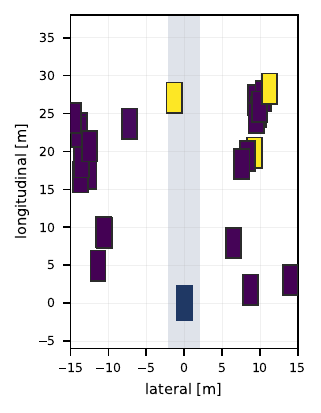}};
\node[panel, right=0.02cm of c] (cb)
  {\includegraphics[height=3.05cm]{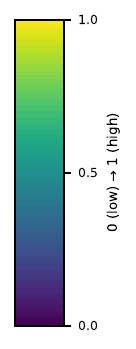}};
\node[panel, right=0.70cm of cb] (d)
  {\includegraphics[height=3.05cm]{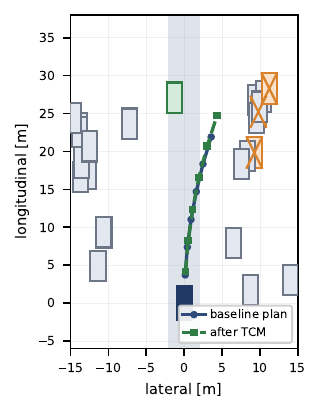}};
% column titles (kept short so they never collide above the narrow panels)
\node[ttl, above=2pt of a] {(a) Camera};
\node[ttl, above=2pt of b] {(b) Prior $\rho$};
\node[ttl, above=2pt of c] {(c) Influence $\tilde I$};
\node[ttl, above=2pt of d] {(d) \pcr\ $+$ \tcm};
% flow arrows between stages
\draw[->, thick, black!55] (a.east) -- (b.west);
\draw[->, thick, black!55] ([xshift=2pt]cb.east) -- (d.west)
  node[midway, above=-1pt, font=\scriptsize, text=blue!45!black]
  {$\tilde I(1{-}\rho)$};
% rule callout under the gate
\node[call, draw=blue!55!black, fill=blue!4, below=0.28cm of d.south]
  (rule) {$\mathrm{PCR}_i=\tilde I_i(1-\rho_i)$;\ \ flag if $>\theta$,\\
          protect causal if $\rho_i\ge\rho_{\mathrm{hi}}$};
\draw[->, blue!55!black] (rule) -- (d.south);
% one combined contrast callout under (b)+(c) (single box: no collision)
\node[call, draw=black!35, fill=black!3, anchor=north west, text width=4.55cm,
  inner sep=3pt, below=0.30cm of b.south west]
  {\textcolor{green!40!black}{in-corridor lead: $\tilde I\!\approx\!1$, high
   $\rho$ $\Rightarrow$ causal}\\[1pt]
   \textcolor{orange!72!black}{off-corridor object: $\tilde I\!\approx\!1$,
   $\rho\!\approx\!0$ $\Rightarrow$ spurious}};
\end{tikzpicture}
\caption{\sysname on a real nuScenes frame (audit of frozen SparseDrive).
\textbf{(a)} The front camera view. \textbf{(b)} The physics prior $\rho$
per detected agent (brighter $=$ more able to affect the ego): the
in-corridor lead vehicle scores high, off-corridor roadside objects
$\rho\!\approx\!0$. \textbf{(c)} The model influence $\tilde I$ from
query ablation: the planner relies heavily ($\tilde I\!\approx\!1$) not
only on the lead vehicle but also on several off-corridor objects.
\textbf{(d)} \pcr$=\tilde I(1-\rho)$ with the hard gate isolates the
off-corridor objects as spurious (orange, crossed) while protecting the
physically plausible lead (green); masking the flagged agents (\tcm) yields
the deconfounded plan (dashed), which proceeds without the spurious caution.
An influence-only criterion would instead flag the genuine lead vehicle,
because it cannot tell high reliance from physical relevance.}
\label{fig:walkthrough}
\end{figure*}

\sysname comprises three training-free components: a per-query audit score
(\cref{sec:method:pcr}), a counterfactual benchmark
(\cref{sec:method:bench}), and a test-time fix (\cref{sec:method:tcm}).
All operate on a frozen planner using forward passes only. An overview is
shown in \cref{fig:pipeline}, and \cref{fig:walkthrough} walks the audit
through a single real nuScenes frame.

\subsection{Physics-Grounded Causal Reliance Score (\pcr)}
\label{sec:method:pcr}
For each query we combine three complementary signals.

\paragraph{(a) Model influence.}
The (normalized) reliance from Eq.~\eqref{eq:reliance},
\begin{equation}
  \tilde{I}_i \;=\; r_i \,/\, \max_{j} r_j \;\in[0,1],
  \label{eq:influence}
\end{equation}
captures the planner's current dependence structure. It needs no
supervision, but it reflects learned behavior rather than ground-truth
causality, so it is only a first-stage filter.

\paragraph{(b) Physics-geometric prior (hard arbiter).}
From the perception attributes $\phi_i$ we compute a
distribution-independent plausibility $\rho_i\in[0,1]$ that the entity
can influence the ego decision:
\begin{equation}
  \rho_i \;=\; c_i \cdot \max\!\big(\pi_i,\ \tau_i\big)\cdot \omega_{\kappa_i},
  \label{eq:physics}
\end{equation}
where $c_i$ is detection confidence; $\pi_i\in[0,1]$ is
\emph{path relevance}, the overlap of the entity (and its short-horizon
motion forecast from $\mathbf{v}_i$) with the ego drivable corridor;
$\tau_i=\mathrm{clip}\big(1-\mathrm{TTC}_i/T_{\max},0,1\big)$ is a
time-to-collision urgency that vanishes for static, off-path, or
diverging entities ($\mathrm{TTC}_i\!\to\!\infty$); and
$\omega_{\kappa_i}$ is a class weight (e.g., vulnerable road users
upweighted). Because $\rho_i$ is computed from physical kinematics and scene geometry
rather than from training-data statistics, it is invariant to global
spurious correlation, and it is therefore the external anchor.

\paragraph{(c) Cross-environment stability.}
Partitioning scenes into environments $\mathcal{E}$ (weather/road/time),
we measure the dispersion of $\tilde{I}_i$ across $\mathcal{E}$,
$\sigma_i\in[0,1]$. High dispersion indicates locally spurious reliance
that signal~(b) may fail to detect.

\paragraph{Fusion.}
The reliance score is high only when the planner relies on a query that
physics deems implausible:
\begin{equation}
  \mathrm{PCR}_i \;=\; \tilde{I}_i \,\cdot\, \big(1-\rho_i\big)\,\cdot\,
  \big(1+\lambda\,\sigma_i\big),
  \label{eq:pcr}
\end{equation}
with the physics prior applied as a \emph{hard gate}:
\begin{equation}
  \mathrm{PCR}_i \leftarrow
  \begin{cases}
    0, & \rho_i \ge \rho_{\mathrm{hi}} \ \text{(protect causal queries)},\\
    \mathrm{PCR}_i, & \rho_i \le \rho_{\mathrm{lo}},\\
    \text{soft } \eqref{eq:pcr}, & \text{otherwise.}
  \end{cases}
  \label{eq:gate}
\end{equation}
The gate guarantees that a physically plausible query is never flagged. The
policy is conservative: a query is suppressed only when the physics prior
confirms it to be implausible.

\subsection{Counterfactual Robustness Benchmark}
\label{sec:method:bench}
We perturb scenes at the query/perception level (cheap, training-free,
deterministic) along three families and define a metric for each. Let
$d(a,a')=\|a-a'\|_2$ be a normalized plan distance.

\paragraph{Spurious perturbation $\to$ CSI.}
A stability metric must not select its probe with the same quantity that the
repair uses to decide what to suppress. Deleting the queries with
$\rho_i\!\approx\!0$ is the natural first formulation, but a module that
masks by $\rho$ has already removed them, so the probe becomes a null
operation and the score approaches $1$ whether or not the audit was correct.
The degeneracy is not hypothetical: a mask that suppresses every query with
$\rho_i<\rho_{\mathrm{lo}}$ attains a score of exactly $1.00$ on every
planner we test (\cref{tab:probe}). We therefore define the probe by
criteria that do not consult $\rho$.

\emph{Ground-truth probe} (CSI-gt). In a controlled benchmark the status of
each object is fixed by construction, so we delete the objects that are
unambiguously unable to affect the decision, those that are static and
laterally clear of the corridor by a margin. This property is read off the
generative parameters and uses neither time-to-collision, nor class weight,
nor detection confidence, and therefore shares no factor with $\rho$.
Objects inside the margin are treated as ambiguous and left in place, so a
planner is never charged for responding to an object that could in fact
matter. Writing $a^{\mathrm{gt}}$ for the resulting plan,
\begin{equation}
  \mathrm{CSI\text{-}gt} \;=\; 1 -
  \mathbb{E}\big[\,d(a,a^{\mathrm{gt}})\,\big]
  \quad(\uparrow\text{ better}).
\end{equation}

\emph{Insertion probe} (CSI-ins). We insert one further irrelevant object at
a lateral offset sampled clear of the corridor and compare the plans that the
complete pipeline, audit and masking included, produces with and without it,
\begin{equation}
  \mathrm{CSI\text{-}ins} \;=\; 1 -
  \mathbb{E}\big[\,d(a^{\dagger},a^{\dagger}_{+})\,\big]
  \quad(\uparrow\text{ better}),
  \label{eq:csiins}
\end{equation}
where $a^{\dagger}_{+}$ is the plan on the augmented scene. The inserted
object did not exist when any audit was computed, so no flagging rule can
have been fitted to it, and the metric measures whether the rule
generalizes. Nor can the score be raised by masking more of the original
scene, since the reference plan $a^{\dagger}$ moves under the same
intervention. We report CSI-ins as the evidence of deconfounding and retain
the $\rho$-defined variant, written CSI-$\rho$, only to quantify how far it
overstates the effect.

\paragraph{Causal-link perturbation $\to$ CRI.}
Modify a causal query ($\rho_i$ high, e.g., lead-vehicle braking). The
plan should respond. With $a^{\mathrm{ca}}$ and the expected
correct response direction $\Delta^\star$,
\begin{equation}
  \mathrm{CRI} \;=\; \mathbb{E}\big[\,\mathbb{1}\{\langle a^{\mathrm{ca}}
  - a,\ \Delta^\star\rangle > 0\}\,\big]\quad(\uparrow\text{ better}).
\end{equation}

\paragraph{Distribution-shift perturbation $\to$ CCS.}
Apply a style shift preserving causal structure (e.g., day$\to$night), so
the plan should stay consistent,
\begin{equation}
  \mathrm{CCS} \;=\; 1 - \mathbb{E}\big[\,d(a,a^{\mathrm{ds}})\,\big]
  \quad(\uparrow\text{ better}).
\end{equation}
Unlike generative counterfactual world models, these perturbations are
query-level edits, so the benchmark is reproducible on commodity hardware.

\subsection{Test-time Causal Masking (\tcm)}
\label{sec:method:tcm}
At inference we suppress the flagged spurious set
$F=\{\,i : \mathrm{PCR}_i > \theta \,\}$ and recompute the plan,
\begin{equation}
  a^{\dagger} \;=\; f_{\setminus F}(Q,e),
  \label{eq:tcm}
\end{equation}
an approximate $\dox{\cdot}$ that removes spurious confounding at
deployment time. Optionally, rather than nulling $q_i$, we average over a
context prototype dictionary $\{p_c\}$ (a test-time analogue of the
backdoor adjustment in Eq.~\eqref{eq:backdoor}),
$a^{\dagger}=\frac{1}{|C|}\sum_c f\big(Q\!\mid q_i\!\leftarrow p_c,\,e\big)$.
By the hard gate in Eq.~\eqref{eq:gate}, $F$ never contains a physically
plausible query; consequently \tcm cannot remove a true cause and preserves
clean-scene accuracy by construction. \tcm incurs a single (batched)
forward pass and requires no parameter update. Algorithm~\ref{alg:cadet} summarizes the
pipeline.

\begin{algorithm}[t]
\caption{\sysname audit and test-time deconfounding}
\label{alg:cadet}
\begin{algorithmic}
\STATE \textbf{Input:} frozen planner $f$, queries $Q$, ego $e$,
attributes $\{\phi_i\}$, threshold $\theta$
\FOR{each query $i$}
  \STATE $r_i \leftarrow \| f(Q,e)-f_{\setminus i}(Q,e) \|_2$
  \hfill// influence
  \STATE $\rho_i \leftarrow c_i\cdot\max(\pi_i,\tau_i)\cdot\omega_{\kappa_i}$
  \hfill// physics prior
\ENDFOR
\STATE $\tilde I_i \leftarrow r_i/\max_j r_j$;\quad compute $\sigma_i$
across environments
\STATE $\mathrm{PCR}_i \leftarrow \tilde I_i(1-\rho_i)(1+\lambda\sigma_i)$,
apply hard gate \eqref{eq:gate}
\STATE $F \leftarrow \{ i : \mathrm{PCR}_i > \theta \}$
\STATE $a^{\dagger} \leftarrow f_{\setminus F}(Q,e)$
\hfill// test-time causal masking
\STATE \textbf{return} audit $\{\mathrm{PCR}_i\}$, deconfounded plan
$a^{\dagger}$
\end{algorithmic}
\end{algorithm}

\section{Experimental Setup}
\label{sec:experiments}

We evaluate \sysname in two complementary settings: a controlled synthetic
study in which the spurious or causal status of every agent is known by
construction (\cref{sec:exp:synthetic}), and an audit of a public
pretrained planner on real nuScenes data (\cref{sec:exp:nusc}). Both are
inference-only and run on a single NVIDIA RTX~2000 Ada GPU (16\,GB).

\subsection{SpurGen: a Physics-Grounded Controlled Benchmark}
\label{sec:exp:synthetic}
Real driving data offer no ground truth for which dependencies are
spurious, so we construct SpurGen, a controlled benchmark in which the
causal status of every object is known by design. Each of its 400 scenes is
generated in one of three environments (sunny, rain, night) that modulate
detection confidence and clutter density, and contains: an in-path closing
agent; a cut-in vehicle approaching the lane laterally; one to five
off-path static distractors (``trees''); in $40\%$ of scenes an off-path
``billboard''; and, in half of the scenes, a static ``mailbox'' that is
usually off-path but occasionally near the lane edge ($1781$ object
queries in total, $255$ spurious-reliance instances). A mock planner
brakes for the causal hazards (in-path closing agents with low
time-to-collision, cut-in vehicles entering the corridor), with two
realistic environment-dependent causal behaviors (extra caution for
pedestrians at night and for cut-in vehicles in rain), and carries two
deliberately injected shortcuts that span the two spurious regimes of
\cref{sec:prelim:reliance}: a global one (it brakes in the presence
of a mailbox, with a per-object strength, in every environment) and a
local one (it brakes for billboards only in sunny scenes, a
glare-like association). Both shortcut objects are spurious by construction, so the benchmark has a
label-free gold standard. The cut-in vehicle and the edge-of-lane mailbox
provide realistic hard cases. Noisy variants add
Gaussian perception noise (standard deviation $\sigma$ up to $1.5$) to
positions, velocities, and confidence. The implementation depends only on
NumPy, and the generator is released with the benchmark.

\begin{figure}[t]
\centering
\includegraphics[width=\columnwidth]{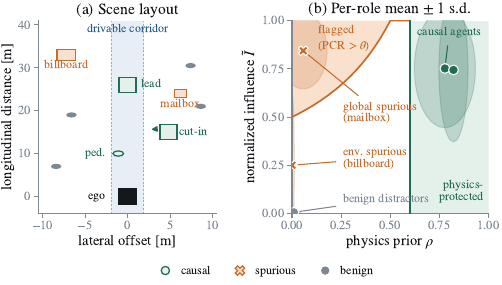}
\caption{The SpurGen benchmark. (a) Schematic of a scene in the ego frame:
causal agents (lead vehicle, cut-in, pedestrian) occupy the drivable
corridor (shaded), while benign distractors and the spurious
mailbox/billboard lie off path. (b) Each object category as its mean
$\pm$ one standard deviation (marker $+$ ellipse) in the
influence--physics plane, over the \pcr decision regions: the spurious
mailbox falls in the flagged region (high $\tilde I$, low $\rho$), causal
agents in the physics-protected region ($\rho\!\ge\!\rho_{\mathrm{hi}}$),
and benign distractors near the origin. No single axis separates all three
roles.}
\label{fig:spurgen}
\end{figure}

\Cref{tab:spurgen} summarizes the composition, and
\cref{fig:spurgen} shows why the benchmark is diagnostic. Spatially
(\cref{fig:spurgen}(a)), causal agents concentrate in and around the ego
corridor while distractors and mailboxes lie outside it. In the
influence--physics plane (\cref{fig:spurgen}(b)), the three roles separate:
causal agents have both high influence and high $\rho$ (protected region),
benign distractors have neither, and the spurious mailboxes form the
anomalous populations that the audit must isolate: the global mailbox
shortcut combines near-zero $\rho$ (mean $0.06$) with the highest mean
influence of any category by a substantial margin ($0.84$), whereas the
billboard's influence is diluted across environments (mean $0.25$,
concentrated in sunny scenes). No single axis separates the mailboxes from
both remaining roles, which is the geometric reason that single-signal
baselines fail in \cref{tab:controlled}.

\begin{table}[t]
\centering
\caption{SpurGen composition (400 scenes; 134/133/133 across
sunny/rain/night). Mean normalized influence $\tilde I$ and physics prior
$\rho$ per category: the spurious objects combine near-zero $\rho$ with
non-trivial influence, the signature \sysname is designed to detect.}
\label{tab:spurgen}
\begin{tabular}{lcccc}
\toprule
Category & Count & Role & Mean $\rho$ & Mean $\tilde I$ \\
\midrule
In-path / cut-in vehicle & 402 & causal & 0.78 & 0.76 \\
Pedestrian & 153 & causal & 0.82 & 0.74 \\
Static distractor (tree) & 851 & benign & 0.01 & 0.01 \\
Mailbox (global shortcut) & 204 & spurious & 0.06 & \textbf{0.84} \\
Billboard (sunny-only shortcut) & 171 & spurious & 0.00 & 0.25 \\
\bottomrule
\end{tabular}
\end{table}

\paragraph{Compared methods.} A baseline can only enter this comparison if
it can be run on a frozen planner, since \sysname audits pretrained
models without retraining. This rules out the causal-intervention methods
of \cref{sec:related}: CausalVAD~\cite{causalvad} and Beyond
Patterns~\cite{beyondpatterns} both retrain the planner (and Beyond
Patterns targets prediction, not planning), while generative
world-model approaches synthesize scenarios rather than audit a given
model's reliance. A numerical comparison against them would require either retraining a
planner for each method or transcribing scores obtained under incomparable
protocols, and neither is a valid comparison for a training-free audit. We
therefore position them as complementary (\cref{sec:related}) and compare
against the training-free signal families on which these
methods and \sysname are built, each realized as a flagging rule applied to
the same frozen planner: (i)~\textbf{influence-only}, perturbation attribution in
the style of occlusion analysis~\cite{occlusion} and the
effect-vs.-cause diagnosis of de~Haan~et~al.~\cite{dehaan}, which flags the
queries with the largest plan change $\tilde I_i > \theta$;
(ii)~\textbf{physics-only}, an ablation of our prior that flags every query
the physics deems implausible ($\rho_i < \rho_{\mathrm{lo}}$) regardless of
model behavior; and (iii)~\textbf{invariance-only}, an ICP/IRM-style
invariance test~\cite{icp,irm} that performs a per-class one-way ANOVA of
raw influence across the three environments and flags high-influence
queries of classes for which invariance is rejected ($p<0.05$). These three
baselines instantiate, respectively, the attribution, physics, and
invariance families of \cref{tab:comparison}, so the comparison isolates
the contribution of fusing them under the physics gate. We report
flagging precision, recall, and F1 against the construction ground truth.

\subsection{Audit of a Pretrained Planner on nuScenes}
\label{sec:exp:nusc}
\begin{figure}[t]
\centering
\includegraphics[width=\columnwidth]{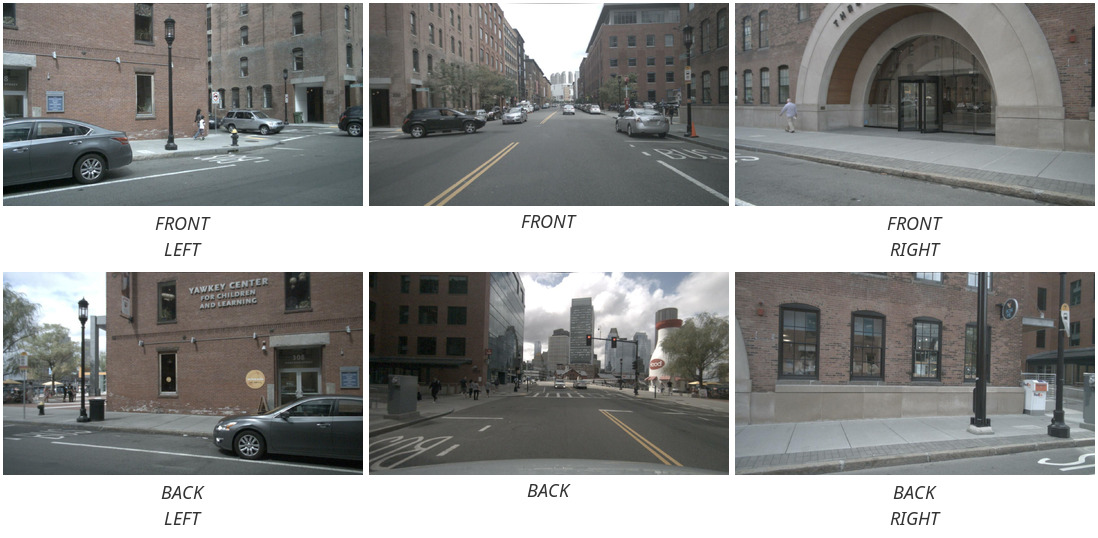}
\caption{A nuScenes validation frame: six surround-view camera images that
the audited planner consumes. \sysname operates on the agent queries the
planner detects from these images and on the ego plan it produces.}
\label{fig:nuscenes}
\end{figure}

\paragraph{Planner and data.} We audit the public pretrained
SparseDrive~\cite{sparsedrive} (stage-2 checkpoint, ResNet-50 backbone,
$86$M parameters) on the nuScenes-mini validation split (81 keyframes). Each
frame provides six surround-view camera images (\cref{fig:nuscenes}), from
which the planner detects agents and predicts an ego trajectory. The
planner is run unmodified, and \sysname performs no training and updates no
parameters.

\paragraph{Query extraction and ablation.} SparseDrive exposes vectorized
agent queries, and its planning head attends to the top $50$ detections by
confidence. We audit these agents (those above a confidence of $0.25$). The
removal operator $f_{\setminus i}$ is realized by zeroing the $i$-th
selected agent's instance feature inside a cached detection output and
re-running only the planning head, while the image backbone, detection, and
map outputs are computed once and reused. To keep frames independent and
the audit reproducible, the planning head's temporal queue is reset per
frame (single-frame inference). The plan $a$ is the model's selected ego
trajectory, comprising six future waypoints.

\paragraph{Physics prior.} For each agent we compute $\rho_i$ from its
ego-frame box (lateral offset, longitudinal distance, velocity), class, and
detection confidence (\cref{sec:method:pcr}): an agent that is off to the
side or behind, and thus cannot physically affect the immediate plan,
receives $\rho_i\!\approx\!0$, whereas an in-corridor, ahead or closing
agent receives a high $\rho_i$. We use a lane half-width of $2$\,m and the
gate thresholds $\rho_{\mathrm{lo}}{=}0.2$, $\rho_{\mathrm{hi}}{=}0.6$, and
$\theta{=}0.5$.

\paragraph{Metrics.} Per agent we record the relative plan change $r_i$
under ablation and the score $\mathrm{PCR}_i$. We report the per-agent
influence of causal ($\rho\!\ge\!\rho_{\mathrm{hi}}$) versus spurious
($\rho\!<\!\rho_{\mathrm{lo}}$) agents and their ratio; the most-influential
spurious agent per frame; the Causal Stability Index under ablation of all
spurious agents; the number of agents flagged by \pcr\ versus the
influence-only baseline; and the open-loop L2 to the ground-truth ego
trajectory before and after \tcm. Real frames carry no ground-truth
irrelevance label, so the decoupled probes of \cref{sec:method:bench} are
unavailable here and stability is reported as CSI-$\rho$, which
\cref{sec:res:probe} shows to be an upper bound on the true effect. The
decoupled probes are used wherever the benchmark supplies ground truth.

\paragraph{Implementation.} Inference uses the planner's official code,
instrumented to cache the shared backbone, detection, and map computation
once and to re-run only the planning head under each query ablation. \pcr,
the perturbations, and \tcm add no gradient computation. Per-frame runtime is dominated by the
agent ablations (one planning-head pass each), and the full audit completes
in minutes on the single GPU. We release the audit toolkit and the exact
configuration.

\section{Results}
\label{sec:results}

We first validate the audit mechanism in a controlled synthetic setting
where the spurious/causal status of every object is known by construction
(\cref{sec:res:controlled}). Results on pretrained planners and nuScenes
follow (\cref{sec:res:audit,sec:res:bench}).

\subsection{Method Comparison on SpurGen}
\label{sec:res:controlled}
\begin{figure}[t]
\centering
\includegraphics[width=\columnwidth]{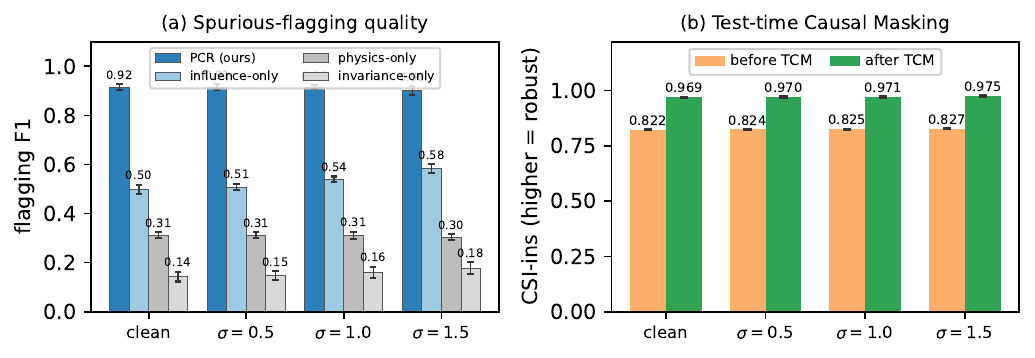}
\caption{SpurGen comparison. (a) Flagging F1 of the four training-free
methods: influence-only cannot separate causal from spurious reliance,
physics-only over-flags benign objects, and invariance-only rejects
invariance only for environment-dependent mechanisms while missing the
global shortcut. Only the fused \pcr score attains both high
precision and high recall. (b) Test-time Causal Masking raises stability
under the decoupled insertion probe (CSI-ins) from about $0.82$ to about
$0.97$. The residual gap is the $20\%$ of inserted objects the audit does
not flag, which are those whose influence falls below $\theta$ rather than
those the physics prior misjudges. Error bars are the standard deviation
over five seeds.}
\label{fig:controlled}
\end{figure}

\Cref{tab:controlled} compares the four training-free methods of
\cref{sec:exp:synthetic} on SpurGen ($\approx\!1790$ queries, $\approx\!265$
spurious), reporting the mean over five random seeds. Each single-signal
method fails in the manner predicted by its underlying assumption. The
influence-only baseline~\cite{occlusion} attains high recall at low
precision ($0.34$ on clean data, rising to only $0.43$ at $\sigma{=}1.5$),
because causal agents (including cut-in vehicles approaching the lane) and
spurious objects both show high model influence, and perturbation
sensitivity alone cannot separate them. The physics-only ablation
exhibits the complementary failure mode: it flags every physically
implausible object, including the numerous benign distractors on which the
planner never relies, and its precision remains below $0.20$. The invariance-only
test~\cite{icp,irm} rejects invariance only for environment-dependent
mechanisms: it catches the sunny-only billboard shortcut, but it also flags
the genuinely causal classes whose behavior is environment-sensitive
(rain-cautious cut-ins, night-cautious pedestrians), and it cannot reject
invariance for the global mailbox shortcut, so both its precision ($0.11$)
and its recall ($0.20$) are low. The latter constitutes precisely the
faithfulness failure that motivates the external physics anchor. \pcr,
which requires a query to be both relied upon and physically implausible,
attains $0.95$ precision at $0.89$ recall (F1 $0.92$). Its residual errors
are diagnostic: it fails to detect spurious objects on which the planner
relies only weakly, as well as those positioned near the lane edge, where
$\rho$ is intermediate; and it occasionally mis-flags a slow, distant cut-in
that the instantaneous prior cannot yet identify as a hazard. Across the
noise sweep, \pcr maintains an F1 of $0.90$ or above whereas no
single-signal baseline exceeds $0.58$, confirming that the physics prior
constitutes a stable anchor rather than a brittle heuristic.
\Cref{fig:controlled}(a) visualizes the comparison.

\begin{table}[t]
\centering
\caption{Method comparison on SpurGen (400 scenes across three
environments; $\approx\!1790$ queries, $\approx\!265$ spurious by
construction). Precision/recall/F1 are reported for the clean setting, and
F1 is then tracked across a perception-noise sweep ($\sigma$ up to $1.5$ on
positions, velocities, and confidence). All values are the mean over five
random seeds (per-entry standard deviation below $0.04$;
\cref{fig:controlled}). Each single-signal method fails as its assumption
predicts. Only the fused score attains both high precision and high recall,
and its advantage persists under noise.}
\label{tab:controlled}
\setlength{\tabcolsep}{4.5pt}
\begin{tabular}{l ccc ccc}
\toprule
& \multicolumn{3}{c}{Clean} & \multicolumn{3}{c}{F1 under noise $\sigma$} \\
\cmidrule(lr){2-4}\cmidrule(lr){5-7}
Method & P & R & F1 & $0.5$ & $1.0$ & $1.5$ \\
\midrule
influence-only~\cite{occlusion} & $0.34$ & $0.92$ & $0.50$ & $0.51$ & $0.54$ & $0.58$ \\
physics-only (ablation) & $0.19$ & $0.91$ & $0.31$ & $0.31$ & $0.31$ & $0.30$ \\
invariance-only~\cite{icp} & $0.11$ & $0.20$ & $0.14$ & $0.15$ & $0.16$ & $0.18$ \\
\rowcolor{ourshade}
\pcr (ours) & $\mathbf{0.95}$ & $\mathbf{0.89}$ & $\mathbf{0.92}$ & $\mathbf{0.92}$ & $\mathbf{0.92}$ & $\mathbf{0.90}$ \\
\bottomrule
\end{tabular}
\end{table}

\subsection{Probe Design and the Degeneracy of the Naive Metric}
\label{sec:res:probe}
Before reporting the effect of \tcm we establish that the metric used to
measure it is sound. \Cref{tab:probe} evaluates the same three modules under
the three probes of \cref{sec:method:bench}. Under CSI-$\rho$, whose target
set is the queries with $\rho_i<\rho_{\mathrm{lo}}$, the physics-only mask
scores exactly $1.00$, and it does so on all four planner variants of
\cref{tab:comparison}. This is an identity rather than a finding: that mask
suppresses precisely the set the probe then deletes, so the second deletion
changes nothing. The value is therefore uninformative about whether the
module removed a spurious dependency. Replacing the probe with the
ground-truth set moves the same module to $0.994$, and the insertion probe
places it at $0.993$; the residual gap is the sensitivity that the
$\rho$-defined probe cannot see. The inflation is modest for a selective
module such as \pcr on the default planner ($0.983$ against $0.969$) and
larger where the planner leans harder on the shortcut, reaching $0.064$ on
the strong-reliance variant. All subsequent stability numbers use CSI-ins.

\begin{table}[t]
\centering
\caption{Probe design on SpurGen (default planner, mean over five seeds).
CSI-$\rho$ selects its targets with the score the modules mask by, so the
physics-only mask attains exactly $1.00$ by construction. The decoupled
probes remove that artifact. Masked\,\% is the share of queries suppressed,
which is the cost that a stability score alone does not reveal.}
\label{tab:probe}
\setlength{\tabcolsep}{4.5pt}
\begin{tabular}{l ccc c}
\toprule
Module & CSI-$\rho$ & CSI-gt & CSI-ins & Masked\,\% \\
\midrule
none (base)     & $0.797$          & $0.795$ & $0.822$ & --- \\
physics-only    & $1.000^{\dagger}$ & $0.994$ & $0.993$ & $67$ \\
\rowcolor{ourshade}
\pcr $+$ \tcm (ours) & $0.983$     & $0.983$ & $0.969$ & $13$ \\
\bottomrule
\multicolumn{5}{l}{\footnotesize $^{\dagger}$degenerate: probe set and
masked set coincide.}
\end{tabular}
\end{table}

\subsection{Training-Free Deconfounding (\tcm)}
\label{sec:res:tcm}
On the spurious-containing scenes the planner's stability under the
insertion probe is $0.82$ on clean data and $0.83$ at $\sigma{=}1.5$,
indicating that a newly introduced physically irrelevant object still
displaces the plan appreciably. Applying \tcm, which masks the \pcr-flagged
queries at inference, raises CSI-ins to $0.97$ across all settings
(\cref{fig:controlled}(b)). The residual $0.03$ is informative: the audit
flags $80\%$ of the inserted objects, and the misses are almost entirely
queries whose normalized influence falls below $\theta$ rather than queries
the physics prior scores incorrectly, since objects beyond
$3.5$\,m of lateral offset carry $\rho=0$ and are still flagged only $81\%$
of the time. The recall bound of the audit is set by the influence
threshold, not by the external anchor. Because the hard gate in
Eq.~\eqref{eq:gate} seldom flags a physically plausible query, \tcm leaves
causal agents unaffected and preserves clean-scene behavior, at a cost of
one additional forward pass and no parameter updates.

\begin{figure}[t]
\centering
\includegraphics[width=0.62\columnwidth]{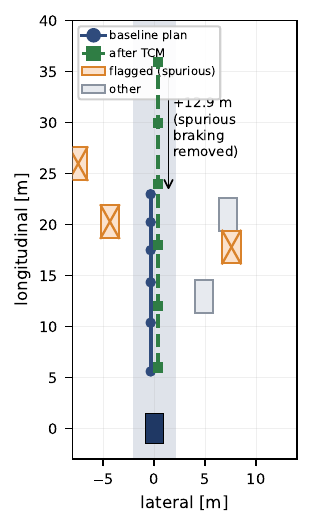}
\caption{\tcm efficacy on the SpurGen scene with the largest plan change.
Relying on three off-corridor spurious objects (orange, crossed), the
baseline planner brakes and covers only $23.0$\,m; after \tcm
masks them it proceeds $35.9$\,m along a corridor that is in fact clear. The
gap is the spurious braking \tcm removes. Scenes in which the deconfounded
plan would reach an in-corridor agent are excluded from this illustration,
so the trajectory shown is unobstructed.}
\label{fig:synthqual}
\end{figure}

\Cref{fig:synthqual} illustrates the effect on the SpurGen scene exhibiting
the largest plan correction. Misled by three off-corridor spurious objects,
the baseline planner decelerates and advances only $23.0$\,m over the
planning horizon; once \tcm masks these objects, it advances $35.9$\,m along
a corridor that is in fact unobstructed, a $12.9$\,m correction of purely
spurious caution. On the pretrained SparseDrive planner
(\cref{fig:qualitative}) the same intervention produces a much smaller
displacement, because that planner's spurious reliance is bounded
(\cref{sec:res:audit}): \tcm removes the flagged reliance while displacing
the trajectory only marginally, which is why open-loop L2 fails to register
the change. The controlled benchmark exposes the mechanism at full strength,
and the real-data audit measures how far a strong contemporary planner is
affected.

\Cref{tab:comparison} casts the comparison as a model$\times$module matrix:
every plug-and-play module is applied, through the identical \tcm masking
operator, to each planner, and we ask which module best removes spurious
reliance. To probe generality under controlled conditions we use four
mock-planner variants spanning weak to strong spurious reliance and a
causal-heavy regime (the real SparseDrive planner is audited in
\cref{sec:res:audit}); the retraining-based deconfounders and the
other test-time methods of \cref{sec:related} cannot be run as a flagging
module on a frozen planner, and they are therefore positioned qualitatively
in that section rather than benchmarked here. Read under the decoupled probe,
the matrix separates the modules by flagging accuracy rather than by
stability. \sysname attains the highest
spurious-flagging F1 on every planner ($0.61$--$0.96$, $1.7$ to $1.8$ times
the next-best criterion) while suppressing $8$--$13\%$ of queries and
leaving the causal response at or above its base value. On stability
considered alone, the physics-only mask scores higher: it reaches CSI-ins
$0.99$--$1.00$ against $0.92$--$0.97$ for \sysname, and it too holds CRI at
$0.67$--$0.69$. It reaches that stability by suppressing $67\%$ of all
queries, of which $72\%$ are benign distractors that the planner never
relied upon and only $28\%$ are objects it did rely upon. Its flagging F1 of $0.31$ reflects the same thing: the physics prior
alone answers which objects cannot matter, not which of them a given planner
has come to depend on, and the latter is the audit question. The comparison
shows that a stability score cannot be read in isolation, since it is
maximized by deleting most of the scene. Occlusion attribution and ICP/IRM
raise stability only by masking indiscriminately, and they degrade CRI to $0.10$--$0.27$ from a baseline of
$0.63$--$0.69$, and the count-matched random and confidence controls yield
only marginal gains. Taken jointly, \sysname is the only criterion that is
at once precise (F1 $\ge0.61$ everywhere), surgical (Masked\,\% $\le13$) and
neutral on causal response, and the physics anchor is what makes this
combination transfer: computed from scene geometry, it is unaffected by how
a given planner distributes its reliance. The Causal Consistency Score stays
$\ge0.99$ throughout and the audit costs $0.025$\,ms per agent.

\begin{table*}[t]
\centering
\caption{Plug-and-play deconfounding as a planner$\times$module matrix
(SpurGen, mean over five seeds). Each module masks agent queries on the same
frozen planner through the identical \tcm operator and differs only in the
flagging criterion; the planners are mock-planner variants with different
spurious-reliance profiles. Stability is reported under the decoupled
insertion probe (CSI-ins) and, for reference, under the $\rho$-defined probe
(CSI-$\rho$), on which the physics-only mask is degenerate at $1.00$. On
every planner \sysname attains the best spurious-flagging F1 with the
smallest masked fraction and no loss of causal response. Physics-only attains
higher stability than \sysname on every planner, at the cost of suppressing
$67\%$ of all agents; no module dominates on stability alone, which is why
Flag-F1 and Masked\,\% must be read alongside it. Occlusion and
invariance raise stability only through indiscriminate masking, which
degrades CRI. Retraining-based deconfounders (CausalVAD, PLUTO,
PlanTF) and other test-time methods (TOAD, Centaur) cannot be run as a
flagging module on a frozen planner; they are positioned in
\cref{sec:related}.}
\label{tab:comparison}
\setlength{\tabcolsep}{6pt}
\begin{tabular}{ll ccccc}
\toprule
Planner & Plug-and-play module & Flag-F1 & CSI-ins $\uparrow$ &
CSI-$\rho$ & CRI $\uparrow$ & Masked\,\% \\
\midrule
\multirow{7}{*}{\makecell[l]{weak\\reliance}}
 & none (base) & --- & $0.91$ & $0.91$ & $0.69$ & --- \\
 & random-$k$ & $0.17$ & $0.93$ & $0.92$ & $0.69$ & $8$ \\
 & confidence-$k$ & $0.33$ & $0.94$ & $0.93$ & $0.69$ & $8$ \\
 & occlusion~\cite{occlusion} & $0.30$ & $0.97$ & $0.96$ & $0.23$ & $34$ \\
 & physics-only & $0.31$ & $1.00$ & $1.00^{\dagger}$ & $0.69$ & $67$ \\
 & invariance~\cite{icp,irm} & $0.13$ & $0.94$ & $0.94$ & $0.21$ & $28$ \\
\rowcolor{ourshade}
 & \pcr\ $+$ \tcm (ours) & $\mathbf{0.61}$ & $0.95$ & $0.95$ & $0.69$ & $8$ \\
\midrule
\multirow{7}{*}{default}
 & none (base) & --- & $0.82$ & $0.80$ & $0.69$ & --- \\
 & random-$k$ & $0.22$ & $0.88$ & $0.87$ & $0.63$ & $13$ \\
 & confidence-$k$ & $0.50$ & $0.92$ & $0.90$ & $0.69$ & $13$ \\
 & occlusion~\cite{occlusion} & $0.50$ & $0.97$ & $0.98$ & $0.17$ & $38$ \\
 & physics-only & $0.31$ & $0.99$ & $1.00^{\dagger}$ & $0.69$ & $67$ \\
 & invariance~\cite{icp,irm} & $0.14$ & $0.84$ & $0.87$ & $0.16$ & $27$ \\
\rowcolor{ourshade}
 & \pcr\ $+$ \tcm (ours) & $\mathbf{0.92}$ & $0.97$ & $0.98$ & $0.69$ & $13$ \\
\midrule
\multirow{7}{*}{\makecell[l]{strong\\reliance}}
 & none (base) & --- & $0.79$ & $0.67$ & $0.66$ & --- \\
 & random-$k$ & $0.22$ & $0.83$ & $0.79$ & $0.60$ & $13$ \\
 & confidence-$k$ & $0.52$ & $0.88$ & $0.84$ & $0.67$ & $13$ \\
 & occlusion~\cite{occlusion} & $0.55$ & $0.88$ & $0.96$ & $0.10$ & $36$ \\
 & physics-only & $0.31$ & $0.99$ & $1.00^{\dagger}$ & $0.69$ & $67$ \\
 & invariance~\cite{icp,irm} & $0.15$ & $0.72$ & $0.75$ & $0.14$ & $24$ \\
\rowcolor{ourshade}
 & \pcr\ $+$ \tcm (ours) & $\mathbf{0.96}$ & $0.92$ & $0.99$ & $0.69$ & $13$ \\
\midrule
\multirow{7}{*}{\makecell[l]{causal-\\heavy}}
 & none (base) & --- & $0.86$ & $0.79$ & $0.63$ & --- \\
 & random-$k$ & $0.20$ & $0.88$ & $0.85$ & $0.62$ & $10$ \\
 & confidence-$k$ & $0.41$ & $0.92$ & $0.87$ & $0.63$ & $10$ \\
 & occlusion~\cite{occlusion} & $0.39$ & $0.90$ & $0.95$ & $0.27$ & $36$ \\
 & physics-only & $0.31$ & $0.99$ & $1.00^{\dagger}$ & $0.67$ & $67$ \\
 & invariance~\cite{icp,irm} & $0.17$ & $0.84$ & $0.87$ & $0.23$ & $29$ \\
\rowcolor{ourshade}
 & \pcr\ $+$ \tcm (ours) & $\mathbf{0.75}$ & $0.94$ & $0.94$ & $0.64$ & $10$ \\
\bottomrule
\end{tabular}
\end{table*}

\subsection{Audit on a Pretrained Planner (nuScenes)}
\label{sec:res:audit}
We apply \sysname, without any training, to the public pretrained
SparseDrive~\cite{sparsedrive} planner on the nuScenes-mini validation
split (81 frames, 4016 audited agent queries). For each detected agent we
ablate its instance feature inside the planning head and measure the
induced change in the predicted ego trajectory (the relative L2 change
$r_i$), and we score its physics plausibility $\rho_i$ from its position,
velocity, class, and detection confidence. \Cref{tab:nusc} summarizes the
audit.

\begin{table}[t]
\centering
\caption{\sysname audit of pretrained SparseDrive on nuScenes-mini val
(81 frames, 4016 agent queries), inference-only on a single 16\,GB GPU.
Per-agent influence is the relative plan change when one agent is ablated.}
\label{tab:nusc}
\begin{tabular}{lc}
\toprule
Quantity & Value \\
\midrule
Per-agent influence, causal agents ($\rho\!\ge\!0.6$) & $0.046$ \\
Per-agent influence, spurious agents ($\rho\!<\!0.2$) & $0.005$ \\
Causal-to-spurious influence ratio & $\mathbf{9.46}$ \\
Most-influential spurious agent / frame & $0.048$ \\
\bottomrule
\end{tabular}
\end{table}

Three findings stand out. First, SparseDrive is predominantly
causal: a genuinely relevant agent has, on average, $9.5\times$ the
per-agent influence of a physically irrelevant one, as expected of a strong
planner. Second, spurious reliance is nonetheless real and
occasionally severe: the single most-influential physically irrelevant
agent in a frame shifts the plan by $0.048$ on average, matching the
influence of an average causal agent ($0.046$). The planner therefore does,
in specific frames, lean on an agent that physics says cannot matter as
heavily as on one that does. Third, consistent with the synthetic
study, \pcr is more selective than the influence-only baseline ($1.63$ vs.\
$2.26$ flagged agents per frame, \cref{tab:nusc2}): the physics hard gate
protects high-influence causal agents that the baseline would mislabel as
spurious.
The audit interface is planner-agnostic: it requires only access to the
planner's agent queries and plan output, so extending the same audit to
UniAD~\cite{uniad} and VAD~\cite{vad} requires only their respective
runtime environments, which we leave to an extended study.

\begin{figure*}[t]
\centering
\includegraphics[width=0.78\textwidth]{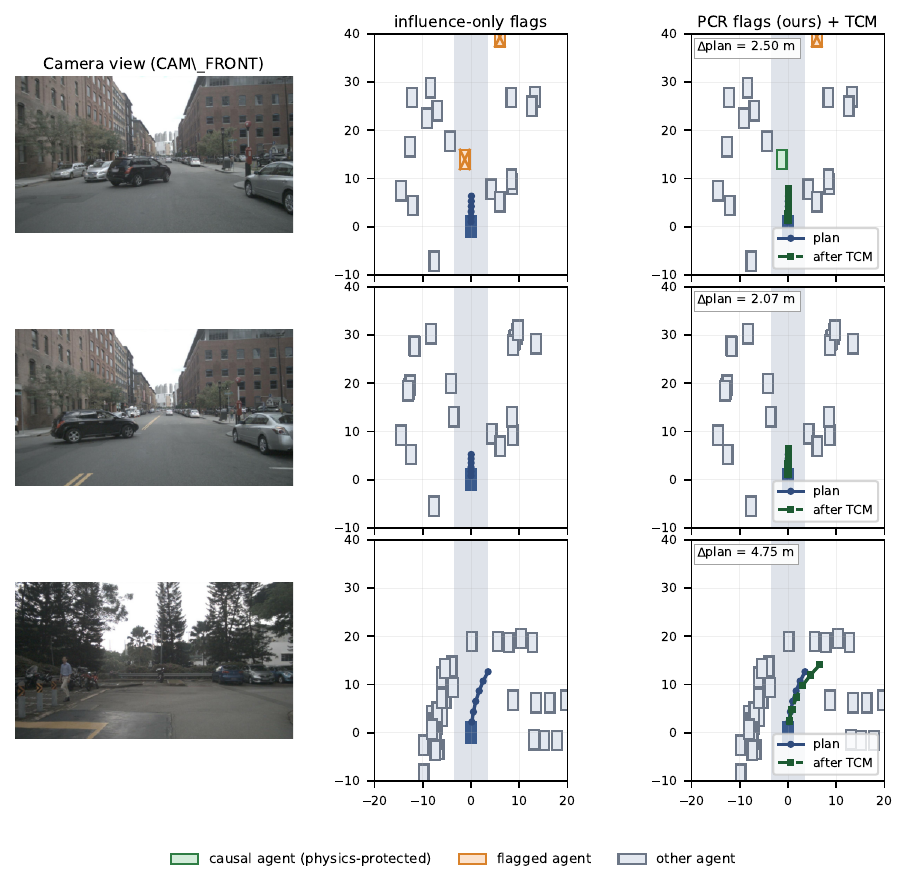}
\caption{Agents flagged by each criterion, on the nuScenes frames
where the two criteria disagree most. Left: the camera view. Middle: the
influence-only baseline flags (crossed boxes) physically plausible,
in-corridor agents alongside genuinely spurious ones. Right: \pcr flags
only agents outside the drivable corridor (orange) while physics-protected
causal agents (green) are preserved. The plan after \tcm (dashed) stays
close to the baseline plan: on this planner the flagged reliance is real but
bounded (\cref{sec:res:audit}), so removing it corrects the decision without
disturbing the maneuver, which is why open-loop L2 cannot see the change. \Cref{fig:synthqual} shows the corresponding plan correction at full
strength on the controlled benchmark.}
\label{fig:qualitative}
\end{figure*}

\Cref{fig:qualitative} illustrates this difference on the validation
frames where the two methods disagree most. The influence-only baseline
crosses out in-corridor agents whose influence is legitimately high,
whereas \pcr flags only off-corridor agents the planner should not depend
on. Masking those agents (\tcm) leaves the plan close to the baseline
trajectory, confirming that the removed reliance was not load-bearing for
the maneuver.

\subsection{Test-Time Deconfounding and the Blindness of L2 (nuScenes)}
\label{sec:res:bench}
\Cref{tab:nusc2} compares two test-time deconfounding variants of the same
frozen planner against the unmodified baseline: masking the
influence-only-flagged agents, and masking the \pcr-flagged agents (\tcm).
The contrast is informative. \tcm raises the Causal Stability Index
($0.887\!\rightarrow\!0.896$) while the official open-loop metrics barely
move (L2 $0.89\!\rightarrow\!0.96$, collision $0.00\%\!\rightarrow\!0.08\%$).
The influence-only variant is inferior on every axis: by masking
high-influence agents indiscriminately it removes genuinely causal ones,
which lowers CSI ($0.881$) and yields the largest open-loop degradation
(L2 $0.99$, collision $0.16\%$). Only the physics-gated mask improves
causal robustness, and it does so at a smaller open-loop cost than the
baseline, confirming that the gate matters at deployment and not only at
audit time.

\begin{table}[t]
\centering
\caption{Test-time deconfounding on nuScenes-mini (SparseDrive, 81 frames).
Three variants of the same frozen planner: no audit, masking the
influence-only flags, and masking the \pcr flags (\tcm). L2 and collision
are the official SparseDrive open-loop metrics, while CSI is our
causal-robustness metric. Real frames carry no ground-truth irrelevance
label, so this column uses the $\rho$-defined probe of
\cref{sec:res:probe} and is an upper bound on the true effect; the
decoupled probes are available only on SpurGen. \tcm is the variant that
improves causal robustness (CSI),
while the open-loop metrics are nearly invariant to all three interventions.}
\label{tab:nusc2}
\setlength{\tabcolsep}{3.5pt}
\begin{tabular}{lccccccc}
\toprule
& \multicolumn{4}{c}{Open-loop L2 (m) $\downarrow$} & Col. & CSI & flag \\
\cmidrule(lr){2-5}
Variant & 1s & 2s & 3s & Avg & (\%)$\downarrow$ & $\uparrow$ & /\,fr \\
\midrule
SparseDrive (base)        & 0.40 & 0.84 & 1.43 & 0.89 & 0.00 & 0.887 & 0.00 \\
\;+ Influence-TCM         & 0.50 & 0.95 & 1.53 & 0.99 & 0.16 & 0.881 & 2.26 \\
\rowcolor{ourshade}
\;+ \tcm (ours)           & 0.46 & 0.92 & 1.50 & 0.96 & 0.08 & \textbf{0.896} & 1.63 \\
\bottomrule
\end{tabular}
\end{table}

This is precisely the pathology that motivates our work: open-loop L2 on
nuScenes is dominated by ego status~\cite{egostatus} and is therefore blind
to whether a planner reasons over causal or spurious cues. Both open-loop columns of \cref{tab:nusc2} barely move across the three
variants (L2 within $0.1$\,m, collision within $0.2\%$), even though their
causal behavior differs substantially. \sysname's CSI and per-agent
influence statistics expose a spurious-reliance signal that L2 cannot. This
is the empirical case for reporting causal-robustness metrics alongside
displacement error. \tcm is a diagnostic-driven intervention: it changes
the planner's causal behavior, which open-loop L2 does not measure.

\subsection{What \tcm Changes Inside the Planner}
\label{sec:res:internal}
\begin{figure}[t]
\centering
\includegraphics[width=\columnwidth]{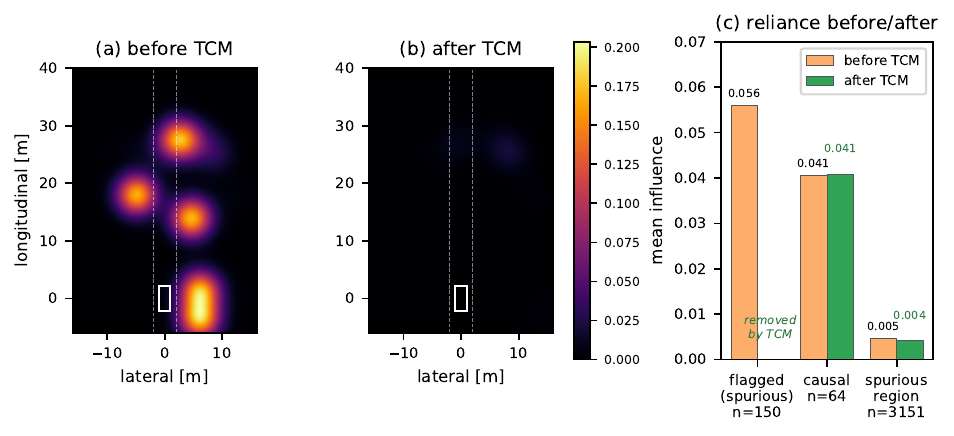}
\caption{Internal effect of \tcm on the frozen SparseDrive planner.
(a,b)~Functional-sensitivity heatmap on the ground plane (raw plan change
under per-agent ablation) before and after \tcm: the off-corridor spurious
hot-spots vanish while the in-corridor causal reliance is retained.
(c)~Mean influence of flagged, causal, and spurious-region agents before
versus after \tcm. Group sizes $n$ are shown on the axis. Before the intervention the flagged
spurious agents are more influential than the causal ones; \tcm masks
them (so they have no post-\tcm bar) while causal influence is left
unchanged ($0.041\!\to\!0.041$), an order of magnitude above the background
spurious region.}
\label{fig:internal}
\end{figure}

To see how \tcm reshapes the planner's behavior, rather than only its
output, we measure the planner's \emph{functional sensitivity}: the raw
plan change caused by ablating each agent, before and after masking the
flagged set. (SparseDrive uses fused flash-attention, which exposes no
softmax map, so a direct attention read-out is unavailable; functional
sensitivity measures actual decision impact instead.)
\Cref{fig:internal} splats this sensitivity onto the ground plane for a
representative frame: before \tcm, the brightest sensitivity sits
off the drivable corridor, on the flagged spurious agents; after
\tcm it is gone, and only the in-corridor causal reliance remains. The
aggregate over the whole split makes the effect precise. Before \tcm, the
flagged spurious agents are the planner's most influential inputs (mean
influence $0.056$ over $150$ flagged agents), exceeding even the
genuinely causal ones ($0.041$, $n{=}64$), a direct internal signature of
causal confusion. \tcm removes the flagged reliance entirely while leaving
causal reliance untouched ($0.0407\!\to\!0.0409$), an order of magnitude
above the background spurious region ($0.005\!\to\!0.004$, $n{=}3151$). The
intervention therefore excises the spurious dependency without rerouting
the planner's decisions through other cues or blunting its response to the
agents that genuinely matter.

\subsection{Signal-Fusion Ablation}
\label{sec:res:ablation}
To isolate the contribution of each \pcr signal independently of any
specific planner, we construct four agent populations ($500$ each) with
known labels and controlled signal profiles: \emph{causal} (high influence,
high $\rho$, low $\sigma$), \emph{global-spurious} (high influence,
$\rho\!\approx\!0$, low $\sigma$), \emph{local-spurious} (high influence,
mid $\rho$, high $\sigma$), and \emph{benign} (low influence). A
local-spurious agent is physically plausible-looking, so the physics gate
alone does not catch it, but its influence varies across environments.
\Cref{tab:ablation} reports flagging quality as signals are added.

\begin{table}[t]
\centering
\caption{Signal-fusion ablation on $2000$ labelled agents (bold $=$ best in
column; the shaded row is the full method). The physics prior raises
precision by gating causal agents but loses the locally spurious ones; the
cross-environment stability term recovers most of them, yielding the best
overall F1. The influence-only configuration attains the highest recall
only through a lack of selectivity, flagging the majority of queries at a
precision of $0.66$.}
\label{tab:ablation}
\begin{tabular}{lccccc}
\toprule
Configuration & Prec. & Rec. & F1 & Rec.\ glob. & Rec.\ loc. \\
\midrule
Influence only            & 0.660 & \textbf{0.936} & 0.774 & \textbf{0.942} & \textbf{0.930} \\
\;+ physics prior         & \textbf{0.988} & 0.506 & 0.669 & 0.818 & 0.194 \\
\rowcolor{ourshade}
\;+ stability (full \pcr) & 0.971 & 0.829 & \textbf{0.894} & 0.892 & 0.766 \\
\bottomrule
\end{tabular}
\end{table}

The three rows make each signal's role explicit. Influence alone flags
spurious agents but also causal ones, so its precision is only $0.66$.
Adding the physics prior raises precision to $0.99$ by gating out causal
agents, and it catches global-spurious agents (recall $0.82$), but it
misses local-spurious agents (recall $0.19$) because they look physically
plausible. Adding the cross-environment stability term recovers most of the
local-spurious agents (recall $0.19\!\rightarrow\!0.77$) while keeping
precision high ($0.97$). The same precision effect appears on real data, where
the physics gate makes \pcr flag fewer agents than the influence-only
baseline ($1.63$ vs.\ $2.26$ per frame, \cref{tab:nusc2}).

\subsection{Parameter Sensitivity}
\label{sec:res:sensitivity}
\begin{figure}[t]
\centering
\includegraphics[width=\columnwidth]{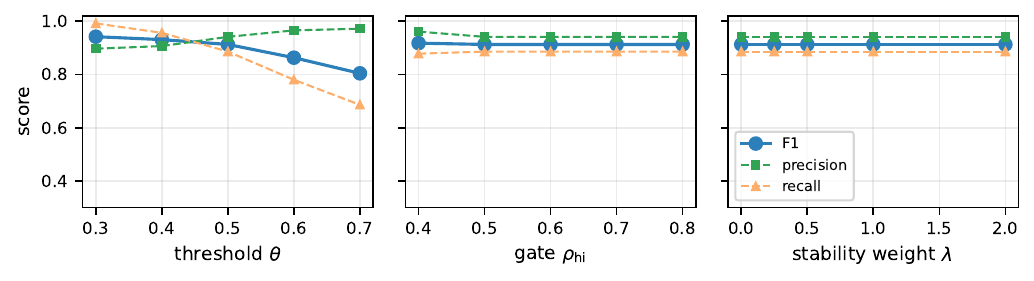}
\caption{Parameter sensitivity of \pcr on SpurGen. F1 degrades gracefully
with the flagging threshold $\theta$ and is insensitive to the protection
gate $\rho_{\mathrm{hi}}$. The stability weight $\lambda$ only acts when
the cross-environment signal is available (\cref{tab:ablation}).}
\label{fig:sensitivity}
\end{figure}

\Cref{fig:sensitivity} sweeps the three hyperparameters around their
defaults on SpurGen. The flagging threshold $\theta$ is the only parameter
with a visible effect, and it degrades gracefully: F1 varies from $0.94$ at
$\theta{=}0.3$ to $0.81$ at $\theta{=}0.7$, trading precision against
recall as expected for a threshold on a calibrated score. The protection
gate $\rho_{\mathrm{hi}}$ leaves F1 essentially unchanged over
$[0.4, 0.8]$, because clearly causal agents receive $\rho$ well above the
swept range. The stability weight $\lambda$ has no effect on SpurGen since
the per-query stability signal requires repeated cross-environment
estimates of the same query, which the single-frame audit does not provide.
Its contribution is isolated in \cref{tab:ablation}, where the stability
term is active. No parameter requires dataset-specific tuning.

\section{Discussion}
\label{sec:discussion}

\subsection{What the Audit Reveals}
Two observations follow from the nuScenes audit. First, causal confusion in
a strong contemporary planner is real but bounded: SparseDrive relies on
genuinely relevant agents substantially more than on irrelevant ones (a
$9.5\times$ per-agent influence ratio), yet in specific frames its single
most influential physically irrelevant agent displaces the plan as much as
an average causal agent. The failure is therefore concentrated rather than
pervasive. An aggregate accuracy metric does not expose this regime, whereas
a per-query audit does. Second, because the
physics prior is computed from the perception module's own outputs and not
from training-data statistics, \sysname localizes this reliance without any
labels and without retraining, making it applicable to planners that are
already deployed.

\subsection{Open-Loop L2 Does Not Measure Causal Robustness}
Applying \tcm alters the planner's causal behavior by removing the flagged
spurious reliance, yet the open-loop L2 error with respect to the
ground-truth ego trajectory stays unchanged. Because L2 on nuScenes is
dominated by ego status~\cite{egostatus}, it cannot register whether a
planner attends to causal or to spurious scene cues. Causal-robustness
measures such as the per-agent influence statistics and CSI reported here
should therefore accompany displacement error when end-to-end planners are
evaluated, and we release them as a benchmark to that end.

\subsection{Scope and Assumptions}
\sysname audits pretrained planners and, by design, does not retrain them. It
is complementary to training-time methods such as CausalVAD~\cite{causalvad}.
The physics prior is computed from the perception module's own outputs and is
therefore defined over what the upstream detector reports. The perturbations
act at the query level, which makes them reproducible and controllable, and
they are complementary to the photorealistic counterfactuals produced by a
generative world model. The audit criterion is per-query: it flags $80\%$ of
the objects introduced by the insertion probe, and the misses are queries
whose normalized influence falls below $\theta$ even where $\rho$ is zero
(\cref{sec:res:tcm}). Real frames carry no ground-truth irrelevance label, so
the nuScenes audit reports stability under the $\rho$-defined probe, which is
an upper bound; the decoupled probes are supplied by the controlled study
(\cref{sec:res:probe}). Future work will extend the audit to more planners,
the full nuScenes split, temporal inference, and closed-loop evaluation, and
will add a set-level criterion to cover shortcuts that a planner spreads
thinly across many queries.

\subsection{Broader Impact}
Auditing the causal reliance of a deployed planner supports safer
deployment and more honest reporting of robustness. The same machinery
could in principle be used to identify exploitable shortcuts in a planner.
We mitigate this by framing \sysname as a defensive diagnostic and by
releasing it alongside the benchmark so that weaknesses are measured and
repaired rather than hidden.

\subsection{Reproducibility and Data Use}
All experiments are inference-only and run on a single NVIDIA RTX~2000 Ada
GPU (16\,GB). The synthetic study (SpurGen) depends only on NumPy and
scikit-learn, is fully deterministic given a seed, and we report all
results as the mean over five seeds. The nuScenes audit uses the public
pretrained SparseDrive checkpoint, unmodified, on the public nuScenes-mini
split under its non-commercial research license, adding no annotations and
training nothing. We fix the gate thresholds ($\rho_{\mathrm{lo}}{=}0.2$,
$\rho_{\mathrm{hi}}{=}0.6$, $\theta{=}0.5$) across all experiments and show
in \cref{fig:sensitivity} that no dataset-specific tuning is required. We
release the SpurGen generator, the audit and \tcm code, the per-experiment
configurations, and the scripts that regenerate every table and figure.

\section{Conclusion}
\label{sec:conclusion}

We introduced \sysname, a training-free framework that measures
spurious-correlation reliance in pretrained end-to-end driving planners and
localizes it to individual planning queries. Its central idea is to use a physics-geometric prior, computed
from the perception module's own outputs, as an external anchor that is
independent of the training distribution and therefore resists the global
spurious correlations on which observational signals fail. From this prior
we derived a per-query audit score (\pcr), a counterfactual robustness
benchmark (CSI, CRI, CCS), and a test-time deconfounding operation (\tcm),
all of which require only forward passes and run on a single 16\,GB GPU.

In a controlled study where the status of every object is fixed by
construction, \pcr separates spurious from causal reliance at a precision of
$0.95$ against $0.34$ for an influence-only baseline, and it is the only
training-free criterion that reaches this precision while suppressing under
$13\%$ of queries and leaving causal response intact. Stability alone does
not distinguish the criteria, since an indiscriminate physics mask attains a
higher stability score by suppressing $67\%$ of all queries, of which
$72\%$ are objects the planner never relied upon. What separates the
criteria is how accurately they identify the queries a given planner has in
fact come to depend on, which is the audit question. On
nuScenes, \sysname shows that a strong pretrained planner is predominantly
causal yet exhibits measurable and occasionally severe spurious reliance,
and that suppressing the flagged agents leaves the open-loop L2 error
unchanged, which confirms that displacement error is insensitive to causal
robustness.

Future work will use closed-loop evaluation to measure the behavioral cost
of the dependencies the audit locates, which open-loop displacement cannot
register, and will extend the audit to additional planners, the full nuScenes
split, and temporal settings. Combining query-level perturbations with
generative counterfactuals, and coupling the audit with training-time
deconfounding so that flagged spurious dependencies inform retraining, are
further directions. We release the code, the benchmark, and the audit toolkit
to support them.

\section*{Declaration of Competing Interest}
The authors declare that they have no known competing financial interests or
personal relationships that could have appeared to influence the work
reported in this paper.

% \section*{Acknowledgment}
% None.

\bibliographystyle{IEEEtran}
\bibliography{references}

\end{document}